\documentclass{article}

\usepackage{arxiv}

\usepackage[utf8]{inputenc} 
\usepackage[T1]{fontenc}    
\usepackage{hyperref}       
\usepackage{url}            
\usepackage{booktabs}       
\usepackage{amsfonts}       
\usepackage{nicefrac}       
\usepackage{microtype}      
\usepackage{lipsum}		
\usepackage{graphicx}

\usepackage{doi}

\usepackage[utf8]{inputenc} 
\usepackage[T1]{fontenc}    
\usepackage{hyperref}       
\usepackage{url}            
\usepackage{booktabs}       
\usepackage{amsfonts}       
\usepackage{nicefrac}       

\usepackage{amsbsy}
\usepackage[ruled]{algorithm2e} 
\usepackage{amssymb}
\usepackage{amsmath,epsfig}
\usepackage{multirow}
\usepackage{graphicx}
\usepackage{subfigure} 
\title{Temporal Alignment Prediction for Few-Shot Video Classification}

\author{ Fei Pan\\
	National Key Lab for Novel Software Technology\\
	Nanjing University\\
	Nanjing, 210023 \\
	\texttt{felix.panf@outlook.com} \\
	\And
	Chunlei Xu\\
	National Key Lab for Novel Software Technology\\
	Nanjing University\\
	Nanjing, 210023 \\
	\texttt{xu.chunlei@outlook.com} \\
	\And
	Jie Guo\\
	National Key Lab for Novel Software Technology\\
	Nanjing University\\
	Nanjing, 210023 \\
	\texttt{guojie@nju.edu.cn} \\
	\And
	Yanwen Guo\\
	National Key Lab for Novel Software Technology\\
	Nanjing University\\
	Nanjing, 210023 \\
	\texttt{ywguo@nju.edu.cn} \\
}

\hypersetup{
pdftitle={Transductive Maximum Margin Classifier for Few-Shot Learning},
pdfsubject={},
pdfauthor={Fei Pan et al.},
pdfkeywords={few-shot, video classification, sequence, similarity learning, temporal alignment prediction},
}

\begin{document}
\maketitle

\begin{abstract}
	The goal of few-shot video classification is to learn a classification model with good generalization ability when trained with only a few labeled videos. However, it is difficult to learn discriminative feature representations for videos in such a setting. In this paper, we propose Temporal Alignment Prediction (TAP) based on sequence similarity learning for few-shot video classification. In order to obtain the similarity of a pair of videos, we predict the alignment scores between all pairs of temporal positions in the two videos with the temporal alignment prediction function. Besides, the inputs to this function are also equipped with the context information in the temporal domain. We evaluate TAP on two video classification benchmarks including Kinetics and Something-Something V2. The experimental results verify the effectiveness of TAP and show its superiority over state-of-the-art methods.
\end{abstract}


\renewcommand{\thefootnote}{}
\footnotetext{Preprint. Work in progress.}
\section{Introduction}
Video action recognition is a hot research topic in computer vision where various methods have been proposed \cite{TSN,Kinetics,twostream,tran2015learning,wang2018non,R21D,xiaolongwang,RSF}. However, they usually focus on learning effective feature representations from a large amount of labeled videos and make predictions based on the learned features. In order to recognize new categories that have never been seen before with deep neural networks, we should collect and label hundreds of videos, which requires a lot of labor cost.

To reduce this cost, few-shot learning which aims to train a model that can generalize well when just a small number of labeled examples are given, has attracted increasing attention in machine learning and computer vision community. Recently, works related to few-shot learning mainly focus on the image domain \cite{Siamese,MatchingNet,ProtoNet,closerlook,LaplacianShot}. Videos contain much more information than still images and we cannot recognize some specific action labels from a single frame in the video. If there are only a few labeled videos per category, it is difficult to learn discriminative feature representations for videos.

Videos usually consist of hundreds of frames with high redundance among consecutive frames. Thus it is necessary to sample a few frames from videos first. If we feed the extracted frames to Convolutional Neural Networks for feature extraction, we get a temporal sequence of feature vectors. Intuitively, we can average over the temporal dimension of the extracted per frame feature vectors. However, such an average pooling way falis to capture the temporal progression of frame-level information. Cao {\it et al.} \cite{TAM} propose Temporal Alignment Module (TAM) for few-shot video classification, which calculates the distance value between videos by a variant of Dynamic Time Warping (DTW). However, the alignments generated by TAM preserve the temporal ordering strictly. In real world applications, local temporal ordering of videos belonging to the same category may also vary.

To address these problems, we propose Temporal Alignment Prediction (TAP) for few-shot video classification that learns the similarities of a query video with respect to the videos in the support set. Specifically, we predict the alignment scores between all pairs of temporal positions in two videos with temporal alignment prediction function. The similarity of the two videos can be obtained by summing the predicted alignment scores multiplied by learned similarities between all pairs of temporal positions. As opposed to TAM, the proposed TAP does not impose the strict order preservation to the alignments. This relieves the limitations of TAM in some real world applications.

Since the temporal alignment prediction function is approximated with neural networks which are deterministic functions, it cannot generate different alignment scores when given the same inputs. Thus, the inputs to the temporal alignment prediction function are also equipped with the context information in the temporal domain. TAP is capable of modeling long-range temporal structures of videos and increasing data utilization efficiency which is significant in the few-shot setting. Besides, our method predicts the alignment scores between all pairs of temporal positions at the same time, such that the computational cost is much smaller than that of TAM which involves a large amount of iterative operations.

Our paper makes three main contributions. First, we propose TAP for few-shot video classification, which increases data utilization efficiency with low computational cost. Second, we predict the alignment scores between all pairs of temporal positions in two videos, which can be used to learn the similarity of the two videos effectively. Third, we evaluate our method on two benchmark datasets for few-shot video classification, and the experimental results show that our method achieves state-of-the-art results. 

\section{Related Work}
\textbf{Few-Shot Learning.} Koch {\it et al.} \cite{Siamese} propose the Siamese Network that measures the similarity of a pair of images. Vinyals {\it et al.} \cite{MatchingNet} propose Matching Network, which produces a weighted nearest neighbor classifier based on the examples in the support set. Snell {\it et al.} \cite{ProtoNet} propose Prototypical Network, which calculates the prototype of each category in the support set and generates distributions over categories for the query examples based on the prototypes. Finn {\it et al.} \cite{MAML} propose MAML, a meta-learning strategy that intends to search for good initialization of the learnable parameters of deep neural networks which adapts to new tasks quickly. Liu {\it et al.} \cite{TPN} propose TPN, that learns to propagate labels from labeled examples in the support set to unlabeled query examples. Ziko {\it et al.} \cite{LaplacianShot} propose a transductive inference algorithm that introduces a unary term based on the class prototypes and a pairwise Laplacian term encouraging the similar query examples to have the same labels.

\noindent\textbf{Video Classification.} Two-stream ConvNet \cite{twostream} utilizes the spatial network to extract the appearance features and the temporal network to leverage temporal information, and makes predictions by fusing the outputs of the two networks at test time. Based on Two-stream ConvNet, Wang {\it et al.} \cite{TSN} propose TSN that uses sparse sampling strategy to model long-range structure in time. C3D \cite{tran2015learning} employs 3D convolutional filters to learn spatio-temporal features for action recognition.  Wang {\it et al.} \cite{wang2018non} propose I3D that utilizes the two-stream 3D CNN with larger size which learns much longer range temporal structures. Du {\it et al.} \cite{R21D} propose a R(2+1)D structure which factorizes the 3D convolutional filters into the spatial and temporal parts. Wang {\it et al.} \cite{xiaolongwang} build a space-time region graphs for capturing the long-range dependencies.

\noindent\textbf{Few-Shot Video Classification.} Zhu {\it et al.} \cite{CMN} propose CMN for few-shot video classification that uses the key-value memory network to store the video level features and classifies test videos by looking up on the memory network. Zhang {\it et al.} \cite{fs-attention} propose ARN consisting of an encoder, a comparator and an attention module to model the temporal structures and solve the distribution shift problems by the attention by alignment mechanism. Cao {\it et al.} \cite{TAM} propose TAM that obtains the distances between videos with DTW and continuously relaxes the discrete operations in DTW to facilitate the end-to-end training of the model.

\noindent\textbf{Sequence Metric Learning.} DTW \cite{DTW1, DTW2} is proposed to measure the similarity between two temporal sequences, which may vary in speed. Cuturi {\it et al.} \cite{SoftDTW} propose a variant of DTW, named Soft-DTW, that employs a soft-minimum operator when calculating the alignment costs. Su {\it et al.} \cite{OPW} propose OPW that transforms the sequences metric learning problem into the Optimal Transport (OT) problem and smoothes the OT problem with temporal regularization terms. Su {\it et al.} \cite{BingSu} propose to learn a ground metric for sequence distance learning.

\begin{figure*}[htbp]
	\centering
	\includegraphics[width=\textwidth]{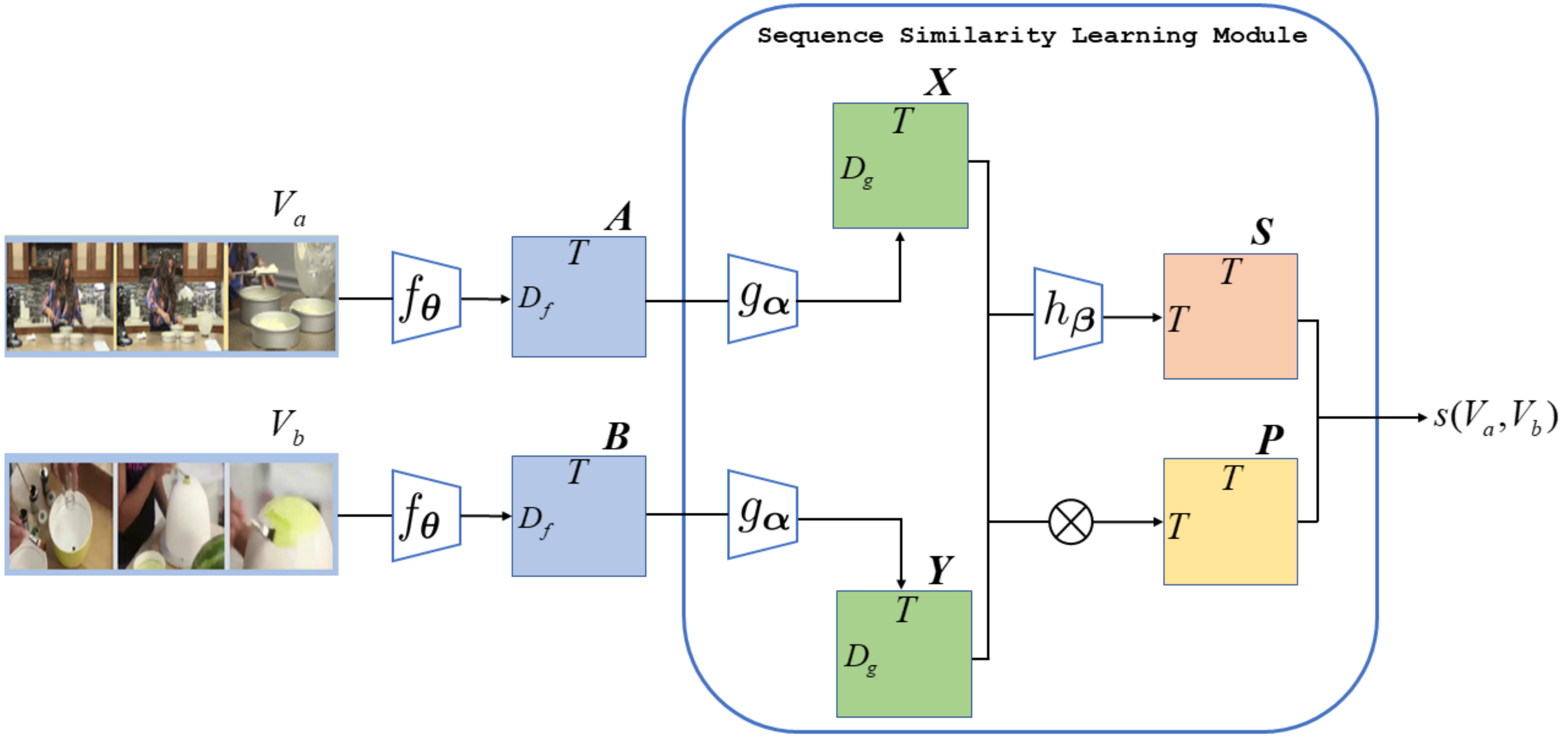}
	\caption{The outline of similarity learning between two videos. We extract features of frames sampled from a pair of videos ($V_a$ and $V_b$) with $f_{\boldsymbol{\theta}}$ and feed them to the Sequence Similarity Learning Module to get the similarity score of the two videos ($s(V_a, V_b)$). The function $g_{\boldsymbol{\alpha}}$ is used to encode the temporal information of sequences. Besides, the function $h_{\boldsymbol{\beta}}$ predicts the alignment scores between all pairs of temporal positions in the two videos. The symbol $\otimes$ denotes the inner product operation. The construction of the matrices $\boldsymbol{A}, \boldsymbol{B}, \boldsymbol{X}, \boldsymbol{Y}, \boldsymbol{S}, \boldsymbol{P}$ will be explained in the main text.} \label{overview}
\end{figure*}
\section{Problem Setting}
We adopt an episodic training strategy \cite{MatchingNet} that learns from tasks over a large amount of episodes for few-shot video classification. In a $N$-way $K$-shot problem, for each episode $\mathcal{T}$, the support set $\mathcal{D}^s$ contains videos with $N$ categories and each category has only $K$ examples. We should classify the videos in the query set $\mathcal{D}^q$ to one of the $N$ categories in the support set. Besides, the episodes are randomly sampled from a meta set. In our setting, there are three meta sets which have no overlapping categories, namely the meta training set, the meta validation set and the meta test set. We train a model using the meta training set, which will be tested in the meta test set. The meta validation set is used for tuning the hyper-parameters.

During training, the optimization is performed based on the prediction results of the query samples in the episode. Specifically, we predict the labels of videos in the query set based on the video and label pairs in the support set. Then we calculate the loss values of all query examples in the episode by a loss function $\mathcal{L}$. In order to search for the optimal model parameterized by $\boldsymbol{w}$, we minimize the averaged loss over multiple episodes:
\begin{equation}\label{eq:q4}
\min_{\boldsymbol{w}} \mathbb{E}_{\mathcal{T} }\left[\frac{1}{Q}\sum_{j=1}^{Q} \mathcal{L}\left(V_j,t_j;\mathcal{D}^s,\boldsymbol{w}\right)\right]\mbox{,}
\end{equation}
where there are totally $Q$ videos in the query set $\mathcal{D}^q$. Besides, $V_j$ and $t_j$ denote a video and the corresponding label in $\mathcal{D}^q$, respectively.

\section{Proposed Method}
Our method can be briefly described as follows. We first sample a few frames from videos and extract per frame features. Then, we learn the similarities of a query video with respect to the videos in the support set. Figure~\ref{overview} shows the outline of the similarity learning between two videos. Moreover, for each category, we compute the mean value of the similarities of the query video with respect to the videos of this category in the support set. Finally, we apply the softmax operator to the mean values to obtain the probability of the query video belonging to all categories and take advantage of the negative log-likelihood loss function for model training.

\subsection{Feature Extraction}
Videos taken from real life may contain actions with multiple stages, each of which lasts for a relatively long time. Thus it is usually necessary to sample a few frames spanning the whole video. We follow the sparse sampling strategy as described in TSN \cite{TSN}: the video is divided into $T$ segments of equal durations and a single frame is randomly sampled from each segment. Sparse sampling makes videos of varying temporal duration to be expressed as sequences of fixed length.  The sparse sampling procedure is denoted as a function $\varphi_T$ in this paper. After a video $V$ is sparsely sampled, a sequence of images $\varphi_T(V)=\{\boldsymbol{x}_{1}, \ldots, \boldsymbol{x}_{T}\}$ is obtained where $\boldsymbol{x}_i$ denotes an image sampled from $V$.

The feature extraction function $f_{\boldsymbol{\theta}}$ parameterized by $\boldsymbol{\theta}$ is used to generate compact feature representations of video frames with the dimension of $D_f$. We leverage $f_{\boldsymbol{\theta}}$ to extract the features of all images in $\varphi_T(V)$ and get a sequence of feature vectors $f_{\theta}(\varphi_T(V))=\{f_{\theta}(\boldsymbol{x}_{1}), \ldots, f_{\theta}(\boldsymbol{x}_{T})\}$.

\subsection{Sequence Similarity Learning}
In the following description, we represent a sequence of vectors as a matrix, e.g. $\boldsymbol{Z} \in \mathbb{R}^{D \times T}$, where $D$ and $T$ denote the dimension of the vector and the length of the sequence, respectively. Besides, $\boldsymbol{Z}_{[i]} \in \mathbb{R}^{D}$ denotes the $i$-th column of the matrix $\boldsymbol{Z}$, $i\in{1,\cdots,T}$. Given two videos $V_a$ and $V_b$, we first sample video frames sparsely using $\varphi_T$, and then feed them into the feature extraction function $f_{\boldsymbol{\theta}}$. Thus, we get:
\begin{equation}
    \boldsymbol{A}=f_{\boldsymbol{\theta}}(\varphi_T(V_a)), \boldsymbol{B}=f_{\boldsymbol{\theta}}(\varphi_T(V_b)), \boldsymbol{A},\boldsymbol{B} \in \mathbb{R}^{D_f \times T}\mbox{.}\label{matrix_ab}
\end{equation}

In this section, we describe the similarity learning between the videos $V_a$ and $V_b$, which is approximated by the similarity of the sequences $\boldsymbol{A}$ and $\boldsymbol{B}$. Previous works usually use average pooling along the temporal dimension of the per frame features to obtain fixed size feature representations for videos \cite{TSN, CMN}. Our method takes advantage of sequence similarity learning for few-shot video classification, and the temporal information of videos is efficiently utilized for label prediction. 

We begin by introducing the DTW algorithm \cite{DTW2}, which can be used to acquire the optimal alignment path between two sequences of vectors $\boldsymbol{E}$ and $\boldsymbol{F}$ ($\boldsymbol{E},\boldsymbol{F}\in\mathbb{R}^{D \times T}$). $\mathcal{I}=[\left(I_{k}^{1}, I_{k}^{2}\right)]_{k=1}^{|\mathcal{I}|}$ is a list of index pairs between the sequences $\boldsymbol{E}$ and $\boldsymbol{F}$, where $I_k^{1}$ and $I_k^{2}$ belong to $\{1, \ldots, T\}$ and $|\mathcal{I}|$ denotes the length of the list $\mathcal{I}$. Then the goal of DTW is to search for the optimal $\mathcal{I}^*$ such that
\begin{equation}\label{eq:warp}
\mathcal{I}^*=\arg\min_{\mathcal{I}} \sum_{k=1}^{|\mathcal{I}|} \|\boldsymbol{E}_{[I_{k}^{1}]} - \boldsymbol{F}_{[I_{k}^{2}]}\|_2^2\mbox{.}
\end{equation}

As shown in Eq.~(\ref{eq:DTW1}), DTW is solved through dynamic programming. $D_{i, j}$ is the distance of the optimal alignment path between the first $i$ columns of the matrix $\boldsymbol{E}$ and the first $j$ columns of the matrix $\boldsymbol{F}$,
\begin{equation}\label{eq:DTW1}
D_{i, j}=\|\boldsymbol{E}_{[i]} - \boldsymbol{F}_{[j]}\|_2^2+\min \left\{D_{i-1, j}, D_{i, j-1}, D_{i-1, j-1}\right\}\mbox{.}
\end{equation}

Then, the distance between the sequences $\boldsymbol{E}$ and $\boldsymbol{F}$ can be obtained by:
\begin{equation}\label{eq:DTW2}
d(\boldsymbol{E}, \boldsymbol{F})=\sum_{k=1}^{|\boldsymbol{I}^*|} \|\boldsymbol{E}_{[I_{k}^{1*}]} - \boldsymbol{F}_{[I_{k}^{2*}]}\|_2^2=D_{T, T}\mbox{.}
\end{equation}

If the columns of the matrices $\boldsymbol{E}$ and $\boldsymbol{F}$ are unit vectors: $\|\boldsymbol{E}_{[i]}\|_2 = \|\boldsymbol{F}_{[i]}\|_2=1, i\in\{1,\cdots,T\}$, Eq. (\ref{eq:DTW2}) can be rewritten as:
\begin{align}
    d(\boldsymbol{E},\boldsymbol{F})&=2|\boldsymbol{I}^*|-2s^{\prime}(\boldsymbol{E},\boldsymbol{F})\mbox{,}\label{eq:DTW3}\\
    s^{\prime}(\boldsymbol{E},\boldsymbol{F})&=\sum_{k=1}^{|\boldsymbol{I}^*|} (\boldsymbol{E}_{[I_{k}^{1*}]})^T \boldsymbol{F}_{[I_{k}^{2*}]}\label{eq:sprime}\mbox{.}
\end{align}

The computation of the similarity value $s^{\prime}(\boldsymbol{E},\boldsymbol{F})$ can be rewritten from Eq.~(\ref{eq:sprime}) to Eq.~(\ref{eq:DTW4}). Here, we introduce a function $\omega$ defined in Eq.~(\ref{eq:P}). $\omega(\boldsymbol{E}_{[i]}, \boldsymbol{F}_{[j]}; \mathcal{I}^*)$ indicates whether the index pair $(i,j)$ belongs to the optimal list $\mathcal{I}^*$. As shown in Eq.~(\ref{eq:DTW4}), the similarity between the sequences $\boldsymbol{E}$ and $\boldsymbol{F}$ can be expressed as the summation of the similarity values $\boldsymbol{E}_{[i]}^T \boldsymbol{F}_{[j]}$ ($i,j\in\{1,\cdots,T\}$) multiplied by the outputs of function $\omega$ evaluated at all pairs of temporal positions. 

\begin{align}
s^{\prime}(\boldsymbol{E}, \boldsymbol{F})&=\sum_{i=1}^{T} \sum_{j=1}^{T} \omega(\boldsymbol{E}_{[i]}, \boldsymbol{F}_{[j]};\mathcal{I}^*) \boldsymbol{E}_{[i]}^T \boldsymbol{F}_{[j]}\mbox{,} \label{eq:DTW4}\\
\omega(\boldsymbol{E}_{[i]}, \boldsymbol{F}_{[j]}; \mathcal{I}^*)&=\left\{\begin{array}{ll}
1, & (i, j) \in \mathcal{I}^* \\
0, & (i, j) \notin \mathcal{I}^*
\end{array}\right.\mbox{.}\label{eq:P}
\end{align}

The sequence similarity learning module used in this paper is comprised of two functions: temporal message passing and temporal alignment prediction. We take advantage of Bidirectional LSTM (BiLSTM) that combines Bidirectional RNNs (BiRNNs) \cite{birnn} with LSTM \cite{lstm} as the temporal message passing function. BiLSTM processes the input vectors with two separate recurrent layers in both the forward and backward directions. The output vectors of the recurrent layers from the two directions are then concatenated as the output of BiLSTM. As a consequence, BiLSTM can access long-range context in two opposite directions.

The temporal message passing function transforms the features of the sampled video frames to another space, where the features at each time step contain the information of the whole sequence. It is denoted as $g_{\boldsymbol{\alpha}}: \mathbb{R}^{D_f \times T} \rightarrow \mathbb{R}^{D_g \times T}$, where $\boldsymbol{\alpha}$ is the learnable parameter and $D_g$ is the dimension of its output features. Thus, we get:
\begin{equation}
    \boldsymbol{X}=g_{\boldsymbol{\alpha}}(\boldsymbol{A}), \boldsymbol{Y}=g_{\boldsymbol{\alpha}}(\boldsymbol{B}), \boldsymbol{X},\boldsymbol{Y} \in \mathbb{R}^{D_g \times T}\mbox{.}
\end{equation}

Afterwards, a pair of columns from $\boldsymbol{X}$ and $\boldsymbol{Y}$ at respective temporal positions are concatenated and passed into the temporal alignment prediction function $h_{\boldsymbol{\beta}}$ parameterized by $\boldsymbol{\beta}$. The function $h_{\boldsymbol{\beta}}: \mathbb{R}^{2D_g} \rightarrow[0,1]$ consists of several stacked fully connected layers followed by a sigmoid activation, which indicates whether a pair of frames are aligned in the feature space. If the value of $h_{\boldsymbol{\beta}}({\rm concat}(\boldsymbol{X}_{[i]}, \boldsymbol{Y}_{[j]}))$ approaches one (zero), it means that the two frames at respective temporal positions $i$ and $j$ in the videos $V_a$ and $V_b$ are (not) aligned. Thus, we obtain the predicted alignment matrix $\boldsymbol{P}\in\mathbb{R}^{T \times T}$ with elements:
\begin{equation}
    \boldsymbol{P}_{ij}=h_{\boldsymbol{\beta}}\left({\rm concat}\left( \boldsymbol{X}_{[i]}, \boldsymbol{Y}_{[j]}\right)\right) ,i,j\in\{1,\cdots,T\}\mbox{.} \label{matrix_M}
\end{equation}

Besides, we define the learned similarity matrix $\boldsymbol{S}\in\mathbb{R}^{T \times T}$ where each element is obtained by the inner product of a pair of the normalized column vectors in the matrices $\boldsymbol{X}$ and $\boldsymbol{Y}$ at respective temporal positions:
\begin{equation}
     \boldsymbol{S}_{ij}=n(\boldsymbol{X}_{[i]})^T n(\boldsymbol{Y}_{[j]}), i,j\in\{1,\cdots,T\}\mbox{,} \label{matrix_C}
\end{equation}
where $n(\cdot)$ is a normalization operator: $n(\boldsymbol{x})=\frac{\boldsymbol{x}}{\|\boldsymbol{x}\|_2}$.

Finally, we replace the function $\omega$ in Eq.~(\ref{eq:DTW4}) which returns discrete values with function $h_{\boldsymbol{\beta}}$ that outputs continuous scores. Thus, the similarity of the videos $V_a$ and $V_b$ is obtained by summing the predicted alignment scores multiplied by the learned similarities between all pairs of temporal positions. It can be equivalently written as:
\begin{align}\label{eq:sim}
s(V_a,V_b)=\sum_{i=1}^{T} \sum_{j=1}^{T} \boldsymbol{P}_{ij}\boldsymbol{S}_{ij}\mbox{.}
\end{align}

The temporal message passing function $g_{\boldsymbol{\alpha}}$ is necessary since the temporal alignment prediction function $h_{\boldsymbol{\beta}}$ is deterministic. This implies that the function $h_{\boldsymbol{\beta}}$ cannot produce different alignment scores when given the same inputs. Thus, the inputs to $h_{\boldsymbol{\beta}}$ should be equipped with the context information in the temporal domain.

Figure~\ref{similaritylearn} illustrates an example of similarity learning between two videos. Figure~\ref{similaritylearn}(b) shows the learned similarity Matrix $\boldsymbol{S}$ obtained by Eq.~(\ref{matrix_C}). Figure~\ref{similaritylearn}(c) and \ref{similaritylearn}(d) demonstrate the predicted alignment matrix $\boldsymbol{P}$ calculated using Eq.~(\ref{matrix_M}) and the corresponding predicted alignments between the two videos, respectively.
From Figure~\ref{similaritylearn}(d), we observe that the temporal ordering is not strictly preserved.

\clearpage
\newcommand{\mysize}{0.45\textwidth}
\begin{figure}[htb]
\centering
\subfigure[two videos of the same category]{
\includegraphics[width=\mysize]{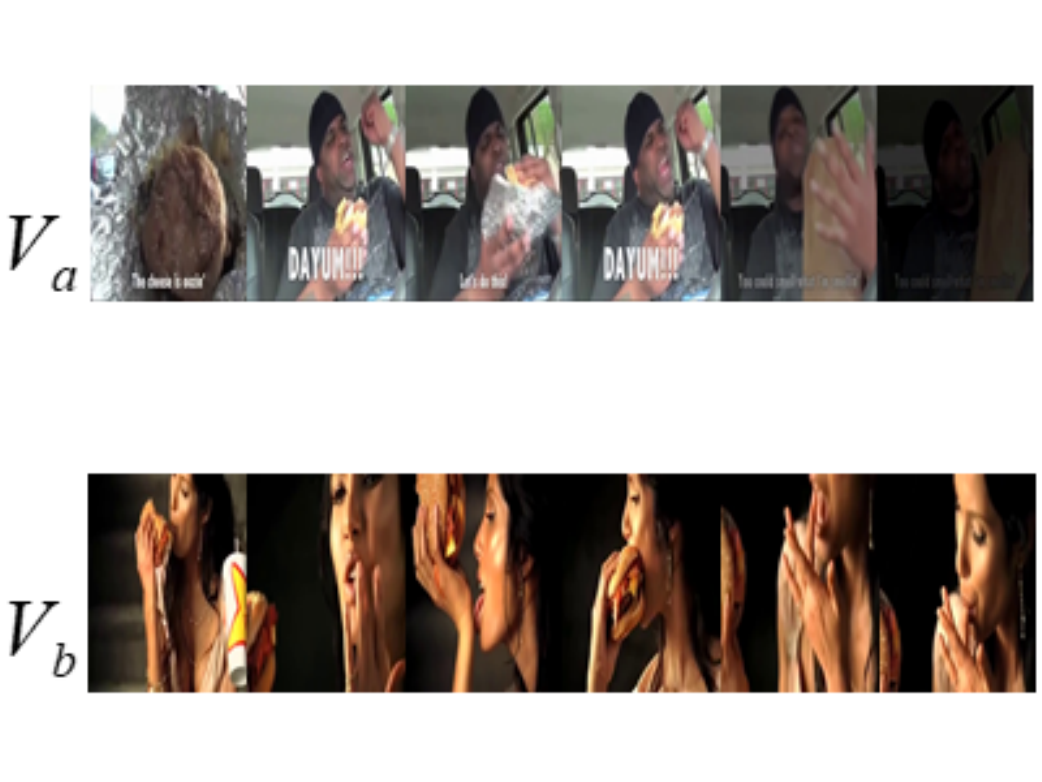}
}
\quad
\subfigure[the learned similarity matrix $\boldsymbol{S}$]{
\includegraphics[width=\mysize]{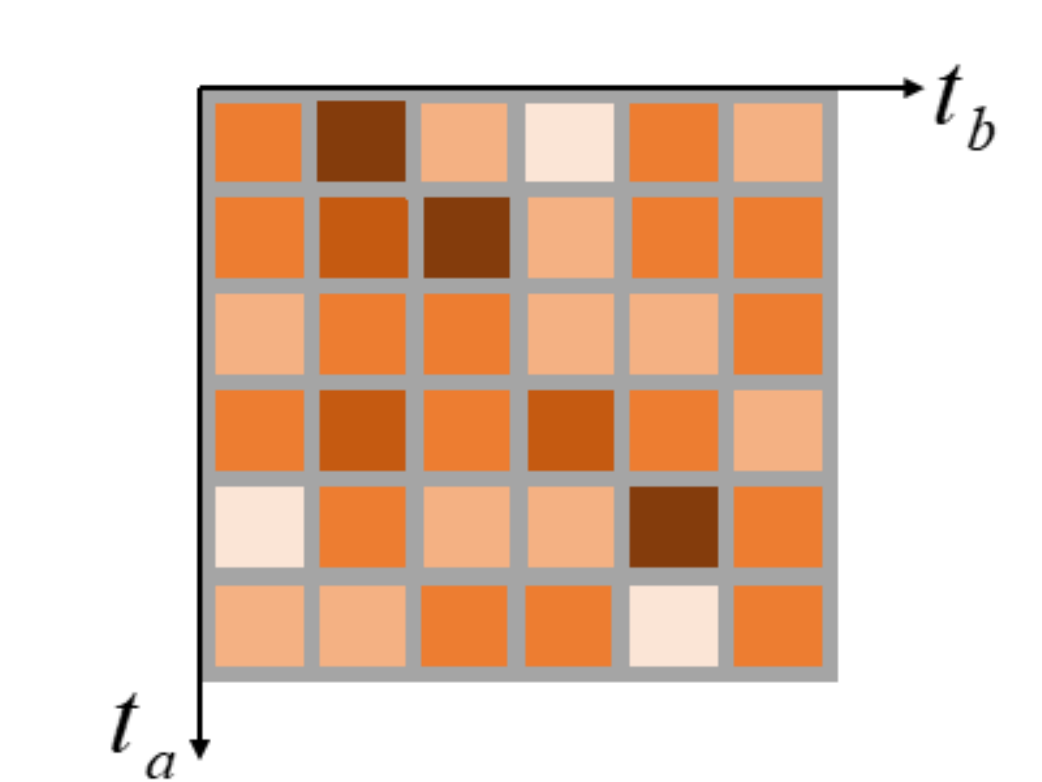}
}
\quad
\subfigure[the predicted alignment matrix $\boldsymbol{P}$]{
\includegraphics[width=\mysize]{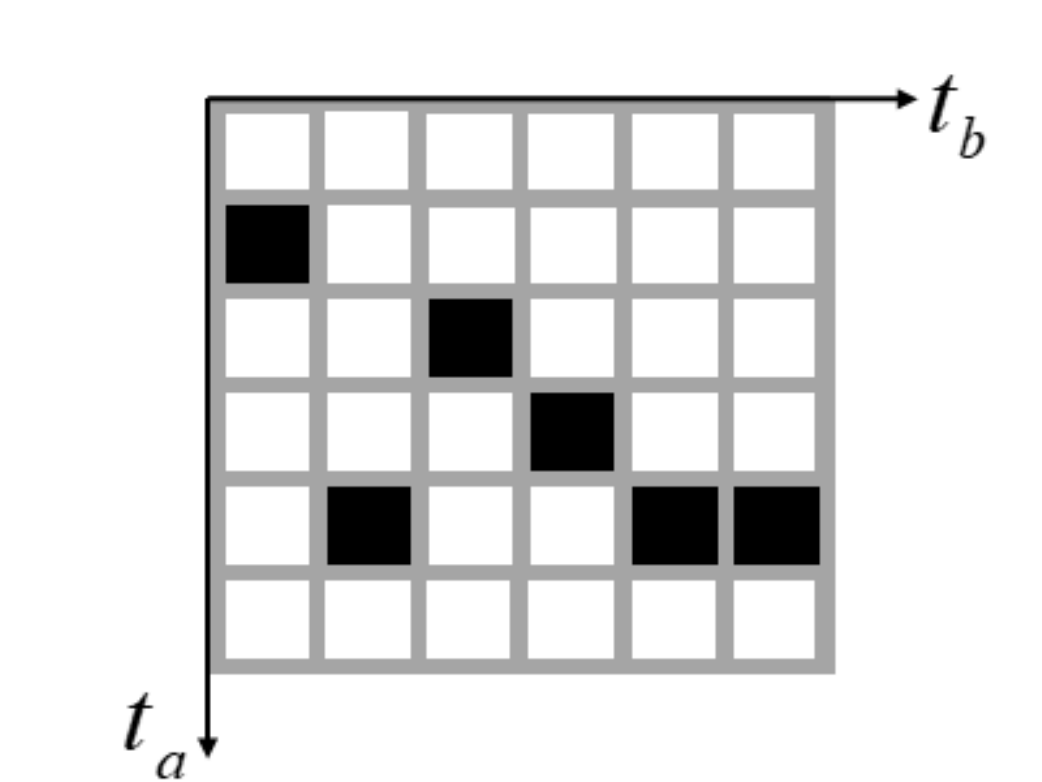}
}
\quad
\subfigure[predicted alignments]{
\includegraphics[width=\mysize]{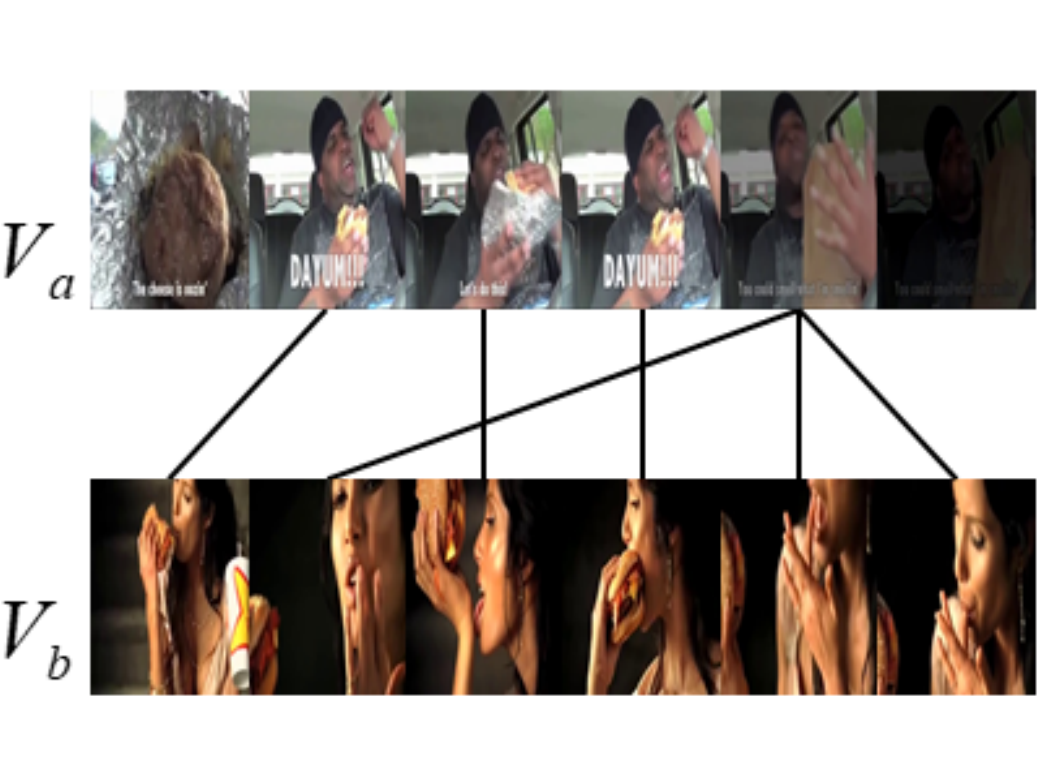}
}
\caption{An example of similarity learning between two videos.}\label{similaritylearn}
\end{figure}

\subsection{Training Objective}
For a $N$-way $K$-shot problem, during training, we are given a query video $V$ with the corresponding label $t$ and the support set consisting of $N\times K$ labeled examples: $\mathcal{D}^s=\{(V_1, t_1), (V_2, t_2),\cdots, (V_{NK}, t_{NK})\}$. Each $V_j$ $(j\in\{1,\cdots,NK\})$ denotes a video and $t_j \in \{1,2,\cdots,N\}$ is the corresponding label. $\mathcal{D}^s_i$ denotes the set of examples labeled with category $i,i\in\{1,\cdots,N\}$. For each category $i$, we compute the mean value of the similarities of the query video $V$ with respect to the videos in $\mathcal{D}^s_i$ by Eq.~(\ref{eq:sim}):
\begin{equation}\label{eq:prototype}
c(V, \mathcal{D}^s_i)=\frac{1}{\left|\mathcal{D}^s_{i}\right|} \sum_{\left(V_{j}, t_{j}\right) \in \mathcal{D}^s_{i}} 
s(V,V_j)\mbox{.}
\end{equation}

We adopt the negative log-likelihood loss function which is given by:
\begin{equation}
    \mathcal{L}(V,t;\mathcal{D}^s,\boldsymbol{w})=-\log \frac{\exp(c(V, \mathcal{D}^s_t))}{\sum_{i=1}^N \exp(c(V, \mathcal{D}^s_i))}\mbox{,}
\end{equation}
where $\boldsymbol{w}$ is the concatenation of the parameters $\boldsymbol{\theta}$, $\boldsymbol{\alpha}$ and $\boldsymbol{\beta}$.

At test time, given the test video $V_{test}$, the optimal label $t_{test}^*$ can be obtained by maximizing the mean values calculated by Eq.~(\ref{eq:prototype}):
\begin{equation}
    t_{test}^*=\underset{i \in \{1, \cdots, N\}}{\arg\max} c(V_{test}, \mathcal{D}^s_i)\mbox{.}
\end{equation}

\section{Experiment}
\subsection{Datasets}
To verify the effectiveness of our method, we experiment on two action recognition benchmarks, namely Kinetics\cite{Kineticsdata} and Something-Something V2 \cite{Something}. The Kinetics dataset contains a wide range of human actions such as shaking hands, robot dancing and riding a bike. The Something-Something V2 dataset is a large collection of videos which show the crowd-source workers performing pre-defined activities. For both datasets, we use 64 categories for meta training, 12 categories for meta validation, and 24 categories for meta testing. 

\subsection{Implementation Details}
For each video, we use the sparse sampling strategy to fetch a few frames. The number of frames sampled from the video is set to 6 in all the experiments. For the sampled frames, we follow the image pre-processing procedures as TSN \cite{TSN}. Specifically, during training, we first resize each frame to $256 \times 256$, and then randomly crop a rectangle with the size of $224 \times 224$. During testing, we use center crop rather than random crop. For Kinetics dataset, we use horizontal flip for data augmentation during training. For Something-Something V2 dataset, since the labels contain left and right information, such as pushing something from left to right and pushing something from right to left, thus horizontal flipping is inapplicable here.

Following the experimental settings of TAM\cite{TAM}, we adopt ResNet-50 as the feature extraction function. The network is initialized with the pretrained models on Imagenet \cite{ImageNet}. The input size, the hidden size and the number of layers of BiLSTM are set to 2048, 1024 and 1, respectively. The output sizes of the stacked fully connected layers in the temporal alignment prediction function are set to 1024, 256, 64, 1, respectively.

During training, we optimize the model using SGD \cite{SGD}, with a start learning rate of 0.001 and we multiply the learning rate by 0.1 every 200 epochs. Besides, weight decay and momentum are set to 0.0001 and 0.9, respectively. We clip gradiens of the learnable parameters $\boldsymbol{\theta}$ and $\boldsymbol{\alpha}$ using the Euclidean norm to avoid explosion of gradients. The maximum norm value is set to 40.

For each episode in a $N$-way $K$-shot task of few-shot video classification, we randomly sample $N$ categories from a meta set and $K+1$ examples for each category. The first $K$ examples are provided for the support set and the last one is supplied to the query set. Thus, each episode contains totally $NK+N$ video examples. We obtain the averaged classification accuracies over 10,000 randomly selected episodes on the meta test set.

\subsection{Results and Comparisons}
We compare our method with existing few-shot video classification methods such as CMN \cite{CMN}, ARN \cite{fs-attention} and TAM \cite{TAM}. Besides, Zhu {\it et al.} \cite{CMN} first average the per frame features to obtain the video-level feature representations and then apply the few-shot image classification methods including Matching Net \cite{MatchingNet} and MAML \cite{MAML} directly. We name the resulting methods MatchingNet+ and MAML+, respectively. Cao {\it et al.} \cite{TAM} also introduce the structure of Baseline++ \cite{closerlook} to several previous algorithms including TSN \cite{TSN}, CMN \cite{CMN} and TRN \cite{TRN}, and obtain the methods TSN++ \cite{TAM}, CMN++ \cite{TAM} and TRN++ \cite{TAM}. The experimental results of these methods are obtained directly from the corresponding papers.
\begin{table*}[htbp]
	\centering
	\begin{tabular}{ccccc}
		\toprule 
		 & \multicolumn{2}{c}{\textbf{Kinetics}} &  \multicolumn{2}{c}{\textbf{Something-Something V2}} \\
		\cline{2-3}
		\cline{4-5}
		Method &  5-way 1-shot & 5-way 5-shot &  5-way 1-shot & 5-way 5-shot\\
		\midrule
		MatchingNet+\cite{CMN} & 53.3 &74.6 & - & - \\
		MAML+\cite{CMN} & 54.2 &75.3 & - & - \\
		CMN\cite{CMN} & 60.5  & 78.9 & - & -\\
		ARN\cite{fs-attention} & 63.7 & 82.4 & - & -\\
		TSN++\cite{TAM} & 64.5 & 77.9 & 33.6 & 43.0\\
		CMN++\cite{TAM} & 65.4 & 78.8 & 34.4 & 43.8\\
		TRN++\cite{TAM} & 68.4 & 82.0 & 38.6 & 48.9\\
		TAM\cite{TAM} & 73.0 & 85.8 & 42.8 & 52.3\\
		\cline{1-5}
		TAP (ours)* & - & - & - & - \\
		\bottomrule
	\end{tabular}
	\caption{Few-shot video classification results. *TAM \cite{TAM} disclosed the split on Something-Something V2 dataset where the number of examples per category is much larger than ours. We will update the experimental results as well as the modified model and experimental settings in the next version.}
	\label{video:result}
\end{table*}

Table~\ref{video:result} shows the comparison of our method to state-of-the-art in $5$-way $1$-shot and $5$-way $5$-shot problems on Kinetics and Something-Something V2 datasets, from which we observe that the performance of our proposed method is state-of-the-art on the two datasets. 

The experimental results also suggest that the methods modeling the temporal structures achieves better classification results than otherwise. Thus sequence modeling is necessary for the few-shot video classification tasks. Effective usage of the temporal information of video enables us to better distinguish different types of videos with a limited number of labeled examples, so as to achieve better classification results. 

In the experiment of CMN\cite{CMN}, it is shown that the classification performance cannot get improved by fine-tuning the backbone networks on the meta training set for few-shot video classification. But in TAM\cite{TAM}, it is found that through appropriate data enhancement and training strategy, the backbone networks can be trained to achieve better results. In our experiments on Kinetics dataset, if the parameters of the feature extraction function (corresponding to the backbone networks in CMN\cite{CMN}) in our model are initialized with ImageNet\cite{ImageNet} pretrained weights and kept fixed during training on the meta training set, the classification accuracy drops by approximately 20\% in 5-way 1-shot tasks on the meta test set .

\subsection{Ablation Study}
In order to evaluate the effectiveness of the proposed method, we remove the sequence similarity learning module from our model. Besides, we use average pooling over the temporal dimension of the features extracted from the sparsely sampled video frames to get the video-level feature representations. Thus, the similarity between a pair of videos is obtained by the inner product of the normalized video-level feature vectors, as apposed to the similarity computation by Eq.~(\ref{eq:sim}) of our method. This procedure results in a model named FewShot-TSN. We report the results of FewShot-TSN compared with TAP in Table~\ref{ablation}.
\begin{table*}[htbp]
	\centering
	\begin{tabular}{ccccc}
		\toprule 
		 & \multicolumn{2}{c}{\textbf{Kinetics}} &  \multicolumn{2}{c}{\textbf{Something-Something V2}} \\
		\cline{2-3}
		\cline{4-5}
		Method &  5-way 1-shot & 5-way 5-shot &  5-way 1-shot & 5-way 5-shot\\
		\midrule
		FewShot-TSN & - & - & - & -\\
		TAP & - & - & - & - \\
		\bottomrule
	\end{tabular}
	\caption{Ablation study of the involvement of the sequence similarity learning module. TAM \cite{TAM} disclosed the split on Something-Something V2 dataset where the number of examples per category is much larger than ours. We will update the experimental results as well as the modified model and experimental settings in the next version.}
	\label{ablation}
\end{table*}

\subsection{Training Time}
\begin{table*}[htbp]
	\centering
	\begin{tabular}{ccc}
		\toprule 
		  &  TAM & Ours\\
		\midrule
		GPU model & TITAN Xp & GeForce 1080 Ti\\
		Number of GPUs & 4 & 4 \\
		Training time & 10 hours & 5 hours\\
		\bottomrule
	\end{tabular}
	\caption{The comparison of the training time with TAM.}
	\label{video:result2}
\end{table*}

We compare the training time of our method with TAM \cite{TAM}, since their performances are comparable. As shown in Table~\ref{video:result2}, the training time of our method is half of TAM. Besides, the performance of GPU model used by TAM is slightly better than that of ours. This indicates that our method is efficient than TAM.

From Eq.~(\ref{eq:DTW1}), it can be seen that there are a large number of iterative operations in solving DTW. Besides, as the number of frames sampled from videos increases, the training time of TAM will increase significantly. However, our method calculates the matching degrees of all pairs of image frames from videos at the same time through temporal alignment prediction. Thus, the computational complexity of our proposed algorithm is much smaller than that of TAM. 

\section{Conclusions}
This paper proposed Temporal Alignment Prediction (TAP) for few-shot video classification. Our method learns the similarity between videos by making full use of the temporal information. Through temporal alignment prediction, the similarity value between a pair of videos is obtained by the summation of the predicted alignment scores multiplied by the learned similarities between all pairs of temporal positions in the two videos. We conducted extensive experiments on two benchmarks including Kinetics and Something-Something V2, where we achieved state-of-the-art results. Our method does not impose strict order preservation to the alignments which may cause failure in some situations. Besides, the computational burden is much smaller compared with TAM.

\bibliographystyle{unsrt}
\bibliography{references}
\end{document}